\documentclass[final]{cvpr}

\usepackage{times}
\usepackage{epsfig}
\usepackage{amsmath}

\usepackage{graphicx} 
\usepackage[dvipsnames]{xcolor} 
\usepackage{booktabs} 
\usepackage[flushleft]{threeparttable} 
\usepackage{multirow} 
\usepackage{soul} 
\usepackage{wrapfig,lipsum,booktabs} 
\usepackage{amssymb}
\usepackage{pifont}
\usepackage{comment}
\newcommand{\PaperTitle}{``BNN - BN = ?": Training Binary Neural Networks without Batch Normalization}

\definecolor{Note_color}{rgb}{1.0, 0.0, 0.0}

\definecolor{Tianlong_color}{rgb}{0.858, 0.188, 0.478}

\usepackage[pagebackref=true,breaklinks=true,colorlinks,bookmarks=false]{hyperref}

\def\cvprPaperID{5} 
\def\confYear{CVPR 2021}

\begin{document}

\title{\PaperTitle}

\author{%
  Tianlong Chen\textsuperscript{1}, Zhenyu Zhang\textsuperscript{2}, Xu Ouyang\textsuperscript{3}, Zechun Liu\textsuperscript{4}, Zhiqiang Shen\textsuperscript{4}, Zhangyang Wang\textsuperscript{1}\\
  \textsuperscript{1}University of Texas at Austin, \textsuperscript{2}University of Science and Technology of China, \\ \textsuperscript{3}Cornell University, \textsuperscript{4}Carnegie Mellon University \\
  \small{\texttt{\{tianlong.chen,atlaswang\}@utexas.edu,zzy19969@mail.ustc.edu.cn,}} \\
  \small{\texttt{xo28@cornell.edu,\{zechunl,zhiqians\}@andrew.cmu.edu}}
}

\maketitle

\begin{abstract}
Batch normalization (BN) is a key facilitator and considered 
essential for state-of-the-art binary neural networks (BNN). However, the BN layer is costly to calculate and is typically implemented with non-binary parameters, leaving a hurdle for the efficient implementation of BNN training. It also introduces undesirable dependence between samples within each batch. Inspired by the latest advance on Batch Normalization Free (BN-Free) training~\cite{brock2021agc}, we extend their framework to training BNNs, and for the first time demonstrate that BNs can be completed removed from BNN training and inference regimes. By plugging in and customizing techniques including adaptive gradient clipping, scale weight standardization, and specialized bottleneck block, a BN-free BNN is capable of maintaining competitive accuracy compared to its BN-based counterpart. Extensive experiments validate the effectiveness of our proposal across diverse BNN backbones and datasets. For example, after removing BNs from the state-of-the-art ReActNets~\cite{liu2020reactnet}, it can still be trained with our proposed methodology to achieve $92.08\%$, $68.34\%$, and $68.0\%$ accuracy on CIFAR-10, CIFAR-100, and ImageNet respectively, with marginal performance drop ($0.23\%\sim0.44\%$ on CIFAR and $1.40\%$ on ImageNet). Codes and pre-trained models are available at: \small{\url{https://github.com/VITA-Group/BNN_NoBN}}.
\vspace{-1em}
\end{abstract}

\section{Introduction}
Despite widespread success~\cite{szegedy2015going,he2016deep,ren2015faster,redmon2016you,long2015fully,chen2017rethinking,pan2016shallow}, state-of-the-art deep networks usually have hundreds of millions of parameters~\cite{brock2021agc,jia2021scaling,pham2020meta}, and suffer from burdensome computational cost. It is questionable how practical they are when it comes to deployment on real-world resource-constrained platforms, e.g., FPGA, ASICs, and mobile devices. Binary neural network (BNN)~\cite{courbariaux2016binarized,courbariaux2015binaryconnect,qin2020binary,rastegari2016xnor,zhou2016dorefa} are therefore proposed for the efficiency purpose. It takes only 1-bit with two discrete values, i.e., $\{-1, 1\}$ to represent networks' weights and activations, leading to significantly accelerated and energy-efficient inference as the 1-bit convolution operation can be efficiently implemented with XNOR and Bitcount operations~\cite{rastegari2016xnor}. 

\begin{figure}[tb]
    \centering
    \includegraphics[width=1\linewidth]{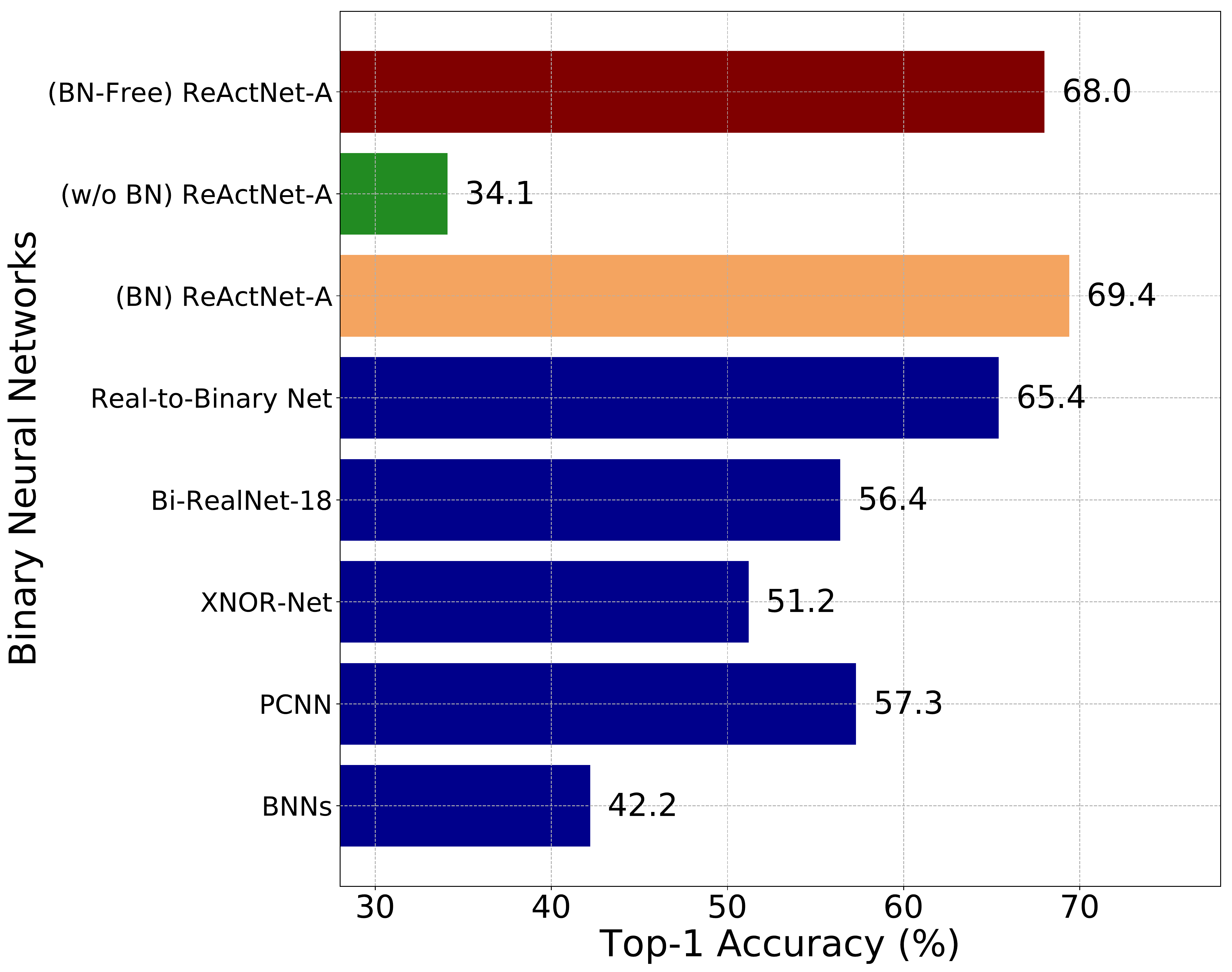}
    \caption{Top-1 accuracies of different binary neural networks (BNN) evaluated on ImageNet. \textcolor{blue}{Blue} bars denote previous BNN methods. The \textcolor{orange}{orange} bar presents the existing state-of-the-art (SOTA) method, i.e., ReActNet~\cite{liu2020reactnet}. The \textcolor{OliveGreen}{green} bar shows the performance of ReActNet if naively dropping Batch Normalization (BN) modules. The \textcolor{Maroon}{red} bar indicates our proposed \textit{\textbf{B}atch Normalization \textbf{F}ree} (BN-Free) binary neural network, which reaches competitive performance compared to its counterpart with BN.}
    \label{fig:teaser}
\end{figure}

Despite these appeals, BNNs are notoriously difficult to train, and undergo performance degradation. Particularly, \cite{santurkar2018does} shows that the Batch Normalization (BN)~\cite{ioffe2015batch} is critical to train BNNs successfully, allowing for stable training under larger learning rates, from both theoretical and empirical perspectives. Unfortunately, the batch normalization implementation~\cite{ioffe2015batch} hinges on high precision values to compute the sum of squares, square-root and reciprocal. Therefore, it comes as no surprise that most  BNNs~\cite{rastegari2016xnor,zhou2016dorefa,liu2018bi,liu2020reactnet} kept BN layers in full precision during training, and some used reduced-precision such as 8-bit~\cite{banner2018scalable}. Although BNs can be absorbed into the BNN weights (e.g., scaling factors) post-training, their presence becomes a bottleneck for BNNs training efficiency on hardware~\cite{wu2018l1}. Moreover, BNs often account for a substantial fraction of run-time, are hard to accelerate~\cite{gitman2017comparison}, and incurs memory overhead~\cite{bulo2018place}. That applies to both inference and the feedforward stage of a training pass. Besides, BN causes discrepant behaviors between the network training and inference stages~\cite{summers2019four}, which may break down the independence assumption between samples within each batch. 



We hence anticipate finding an alternative to eliminate the unwanted properties of BNs in BNNs, while maintaining competitive performance. Motivated by the recent advance~\cite{brock2021characterizing,brock2021agc}, we propose \text{B}atch Normalization \text{F}ree (BN-Free) binary neural networks. Specifically, we leverage the adaptive gradient clipping to constraining BNN's gradient distribution and mitigate gradient explosion due to removing BNs~\cite{santurkar2018does}. Then, the scaled weight standardization and specialized bottleneck block~\cite{brock2021agc} are integrated for preserving the variances and preventing the mean shifts of activations. Our contributions are outlined as follows:
\begin{itemize}
\item We provide the first proof-of-concept study that general BNNs can be successfully trained without BNs but maintain competitive performance. 
\item We introduce adaptive gradient clipping, scaled weight standardization, and specialized block to BNNs, and show these techniques can be easily plugged in various BNN backbones to make them BN-Free.
\item Comprehensive experiments validate the effectiveness of our proposed mechanisms. For example, BN-Free ReActNet achieves $92.08\%$, $68.34\%$, and $68.0\%$ accuracy on CIFAR-10, CIFAR-100, and ImageNet respectively, with only marginal performance drops compared to state-of-the-arts.
\end{itemize}

\section{Related Work}
\paragraph{Binary neural networks.} Numerous model compression and acceleration algorithms have been proposed to reduce the latency of models while maintaining comparable accuracy performance. General model compression approaches fall under multiple forms~\cite{cheng2017survey}: pruning \cite{han2015learning,wen2016learning}, quantization \cite{wu2016quantized,shen2020fractional,fu2020fractrain}, knowledge distillation \cite{hinton2015distilling,mishra2017apprentice}, as well as their compositions \cite{wang2019e2,you2020shiftaddnet,zhao2020smartexchange}.

A Binary Neural Network (BNN)~\cite{courbariaux2015binaryconnect,courbariaux2016binarized,kim2016bitwise,zhou2016dorefa,rastegari2016xnor,courbariaux2016binarized,li2017performance,polino2018model,mishra2017apprentice,hou2016loss,zhou2017incremental,liu2018bi,helwegen2019latent,liu2019circulant,cai2017deep,shen2019searching,han2020training,xu2021learning} represents the most extreme form of model quantization as it quantizes weights in convolution layers to only $1$ bit, enjoying great speed-up compared with its full-precision counterpart. \cite{qin2020binary} roughly divides previous BNN literature into two categories: ($i$) native BNN~\cite{courbariaux2015binaryconnect,courbariaux2016binarized,kim2016bitwise} which directly applies binarization to a full-precision model by a pre-defined binarization function. Straight-through estimator (STE)~\cite{hinton2012,bengio2013estimating} is usually adopted to enable the back-propagation in binarized models~\cite{courbariaux2015binaryconnect}. ($ii$) optimization-based BNNs techniques, including minimizing the quantization error \cite{zhou2016dorefa,rastegari2016xnor,courbariaux2016binarized,li2017performance}, improving the network loss function \cite{polino2018model,mishra2017apprentice,hou2016loss,zhou2017incremental}, and reducing the gradient error \cite{liu2018bi,helwegen2019latent,liu2019circulant,cai2017deep}.

However, such aggressive quantization usually results in severe accuracy decline. To tackle this limitation, \cite{courbariaux2016binarized} proposes an end-to-end gradient back-propagation framework for training the discrete binary weights and activations, establishing great successes on small datasets, such as MNIST~\cite{umuroglu2017finn} and CIFAR10~\cite{wang2019learning}, while still has unsatisfactory performance on large datasets like ImageNet \cite{liu2020reactnet}. Follow-up researches~\cite{bulat2019xnor,gu2019projection,Martinez2020Training,liu2018bi,liu2020reactnet} devote themselves to build state-of-the-art (SOTA) accuracies on ImageNet. Among these works, ReActNet~\cite{liu2020reactnet} proposes generalized activation functions and a distributional loss, reaching the superior performance which reduces the gap to its full-precision counterpart within $3.0\%$ accuracy on ImageNet. Note that, all mentioned SOTA BNNs are not sustained without batch normalization.

\paragraph{Batch normalization and normalization-free networks.} Batch normalization (BN)~\cite{ioffe2015batch} is a well-known and widely used technique to stabilize model training. It also plays a critical role in the BNN training, as evidenced by~\cite{santurkar2018does}. However, bath normalization is an expensive computational primitive~\cite{gitman2017comparison}, and its inefficiency is further amplified in low bits precision context which hinders the deployment of BNN to resource-limited hardware~\cite{wu2018l1}.

To seek a simple and effective alternative for batch normalization, various studies~\cite{zhang2019fixup,bachlechner2020rezero,hanin2018start,qiao2019micro,brock2021characterizing,brock2021agc} are proposed. \cite{zhang2019fixup} introduces an initialization and rescaling rule (i.e., fixed-update initialization) to stabilizes the training of very deep models in place of BN. \cite{bachlechner2020rezero,hanin2018start} share similar observations that appropriately initializing weights and scaling residual modules benefit avoiding the gradient exploding and vanishing, leading to a stabilized training. Another promising substitution is weight standardization~\cite{qiao2019micro}, which subtracts the mean from weights and divides weights by their standard deviation. \cite{brock2021characterizing} proposes a modified variant, i.e., scaled weight standardization, to suppresses the quickly enlarging of the mean in hidden activations. Recently, \cite{brock2021agc} proposes Adaptive Gradient Clipping (AGC) to enable the larger batch size training of normalization-free networks, and to overcome the instabilities from eliminating BN.

\begin{figure*}[t] 
\centering
\includegraphics[width=1\linewidth]{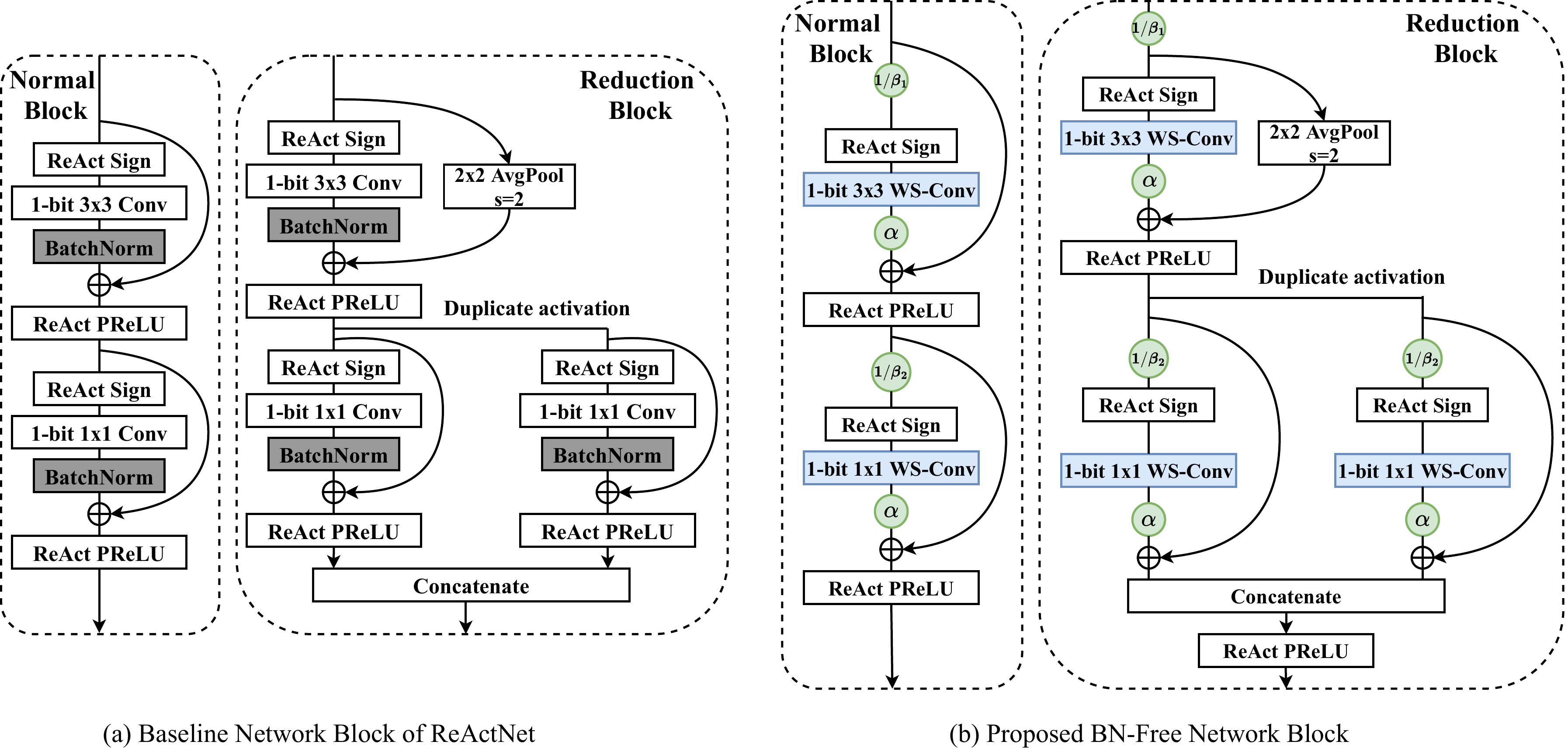}
\caption{The architecture overview of baseline network block (a) and proposed BN-Free network block (b). The baseline network blocks are inherited from the recent state-of-the-art (SOTA) BNN framework, i.e., ReActNet~\cite{liu2020reactnet}, which are modified from MobileNetV1~\cite{howard2017mobilenets} and have the same configuration of channel and layer numbers. For the reduction block, \cite{liu2020reactnet} duplicates the input activation and concatenate the outputs to increase the channel number, which is also maintained in our proposed BN-Free network block. The most important thing is that all original Batch Normalization modules are \textit{removed}, replaced by scaling factors (e.g., $\alpha$, $1/\beta_1$, $1/\beta_2$) and adjusted convolutional layers with scaled weight standardization (i.e., WS-Conv).}
\label{fig:bfblock}
\end{figure*}

\section{Technical Approach}
In this section, we present the detailed normalization-free methodologies for binary neural networks (BNN) in Section~\ref{sec:method_nf} and adopt BNN backbone architecture in Section~\ref{sec:bnn_arch}. Before that, we briefly list the main Batch Normalization benefits from previous literature.

\paragraph{Understanding Batch Normalization.} The Batch Normalization (BN) can ($i$) reduces the scale of hidden activations on the residual branches~\cite{de2020batch,balduzzi2017shattered,hanin2018start,yang2019mean}, and maintains well-behaved gradients early in training; ($ii$) eliminates mean-shift by enforcing the mean activation of each channel to zero across the current training batch~\cite{de2020batch,DBLP,brock2021characterizing}; ($iii$) serves an implicit regularization~\cite{luo2018towards} and enhances the models' generalization~\cite{hoffer2017train}; ($iv$) enables large-batch training~\cite{goyal2017accurate} and smoothens the loss landscapes~\cite{santurkar2018does}. 

Removing batch normalization directly usually leads to an inferior performance~\cite{ioffe2015batch,brock2021agc}. It is further aggravated in training the binary neural network, due to its challenge regime with discrete values of variables~\cite{santurkar2018does}. Particularly, \cite{santurkar2018does} provides both theoretical and empirical analyses to demonstrate the critical role of BN is to alleviate exploding gradients in the case of binary neural networks, which motivates us the introduce adaptive gradient clipping to establish the framework of BN-Free BNN.

\subsection{Normalization-free Training Methodology} \label{sec:method_nf}
\paragraph{Adaptive gradient clipping (AGC).} Gradient clipping is typically adopted to constrain the norm of gradients~\cite{pascanu2013difficulty}, leading to stabilized training~\cite{merity2017regularizing}. Recently \cite{brock2021agc} proposes adaptive gradient clipping (AGC) to ameliorate the NF-ResNets~\cite{brock2021characterizing}'s performance, which clips gradients based on the unit-wise ratios of gradient norms to parameter norms. It can be described as follows:
\begin{equation}
G_i^l\rightarrow\left\{
\begin{aligned}
& \lambda\frac{\|W_i^l\|^*_F}{\|G_i^l\|_F}G^l_i & \text{if } \frac{\|G_i^l\|_F}{\|W_i^l\|^*_F} > \lambda \\
& G_i^l & \text{otherwise.}
\end{aligned}
\right.
\end{equation}
Where $G_i^l$ denotes the $i_{\mathrm{th}}$ row of gradient matrix $G^l$; similarly, $W_i^l$ is the $i_{\mathrm{th}}$ row of weight matrix $W^l$; $l$ is the layer index of the considered network; $\|W_i^l\|^*_F=\mathrm{max}\{\|W_i\|_F,\epsilon\}$, $\epsilon=10^{-3}$ and $\|\cdot\|_F$ is the Frobenius norm.

The clipping threshold $\lambda$ is a crucial hyperparameter, which is usually tuned by a grid search. Equipped with AGC, BNN training tends to have a constrained gradient distribution as evidenced in Figure~\ref{fig:grad}, avoiding the gradient explosion issue.

\paragraph{Scaled weight standardization.} To deal with the mean-shift in the hidden activation distributions caused by removing BN, we also introduced the Scaled Weight Standardization from \cite{brock2021characterizing}. Specifically, we modify all convolutional layers in BNN backbones as follows:

\begin{equation}
\hat{W}_{i,j}=\gamma \cdot \frac{W_{i,j}-\mu_i}{\sqrt{N}\sigma_i}
\end{equation}
where $\mu_i=(1/N)\Sigma_j W_{i,j}$, $\sigma^2_i = (1/N)\Sigma_j(W_{i,j}-\mu_i)^2$, $N$ is the fan-in, and $\hat{W}_{i,j}$ is the corresponding standardized weights. $\gamma$ is a fixed scalar for variance preserving, and has diverse values for different adopted activation functions~\cite{brock2021characterizing}. For example, $\gamma=\sqrt{2/(1-(1/\pi))}$ for the ReLU activation function~\cite{arpit2016normalization}. We name the modified convolutional layer as \textit{WS-Conv} for simplicity. Note that, such WS-Conv has consistent performance between training and inference, mitigating the discrepancy behaviour of the batch normalization~\cite{brock2021agc} and leading to a hardware-friendly implementation of BN-Free binary neural networks.

\paragraph{Specialized bottleneck block.} For the batch normalization benefits preserving purpose, we inherit the specialized bottlenecks block from \cite{brock2021characterizing,brock2021agc} that applies input/output normalization with hand-crafted scaling factor (e.g., $\alpha,\beta$). As shown in Figure~\ref{fig:bfblock}, we utilize $x_{i_0}$ and $x_{i_1}$ to present the input of the $i_{\mathrm{th}}$ BN-Free block and activation after the ReAct PReLU function. In order to normalize the input variance, $\beta_1 = \sqrt{\mathrm{Var}(x_{i_0})}$ is adopted before the 3x3 WS-Conv operation. We then multiply it with a scalar $\alpha$ and feed it to the ReAct PReLU. Similarly, we divide the obtain activation $x_{i_1}$ with $\beta_2 = \sqrt{\mathrm{Var}(x_{i_1})}$ and multiply it with $\alpha$. Blessed by the variance preserving design~\cite{brock2021characterizing,brock2021agc}, the output variance of the $i_{\mathrm{th}}$ BN-Free block is $\mathrm{Var}(x_{i_1})+\alpha^2$. Note that, $\beta_1$ and $\beta_2$ are usually the expected empirical standard deviation of the corresponding activation at initialization~\cite{brock2021agc}.  

\subsection{The Backbone Architecture of BNN} \label{sec:bnn_arch}
\paragraph{Generalized activation functions.} \cite{rastegari2016xnor,xu2019accurate,bulat2019xnor,liu2020reactnet} advocate that enforcing binary neural networks to learn similar distribution as full-precision (i.e., real-valued or 32 bits) networks plays a significant role in the final achievable performance of BNN. Specifically, XNOR-Net~\cite{rastegari2016xnor} pursues close logits distribution as real-valued ones by calculating analytical real-valued scaling factors and multiplying them with the activations. \cite{xu2019accurate,bulat2019xnor} introduce further improvements by learning these factors through back-propagation. ReActNet~\cite{liu2020reactnet} explores an orthogonal perspective that mimics the activation distribution from a pre-trained full precision model. However, it is challenging for binary neural networks with a highly limited capacity to learn appropriate activation distribution, since even small variations to their activation distribution can substantially affect the feature representations in BNNs~\cite{liu2020reactnet}.

To tackle this issue, \cite{liu2020reactnet} proposes the generalized activation functions with learnable parameters, for $\mathrm{sign}$ and PReLU \cite{he2015delving} functions, which are termed as RSign and RPReLU respectively. Such learnable parameters enable the adaptive reshape and shift of BNNs' activation to match the desired distributions. Following \cite{liu2020reactnet}'s definition, we introduce adopted activation functions.

\begin{equation}
\mathrm{(RSign)}\ \  x_i^{b}=h(x_i^{r})=\left\{
\begin{aligned}
+1 & , \text{if } x_i^{r} > \alpha_i \\
-1 & , \text{if } x_i^{r} \leq \alpha_i
\end{aligned}
\right.
\end{equation}
where $x_i^{r}$ is full-precision input of the RSign function $h(\cdot)$ on the $i$th channel, $x_i^{b}$ is the binary output and $\alpha_i$ is a learnable coefficient controlling the threshold. The superscripts $b$ and $r$ above $x_i$ denote the corresponding binary and full-precision values.

\begin{equation}
\mathrm{(RPReLU)}\ \ f(x_i)=\left\{
\begin{aligned}
x_i- \gamma_i + \zeta_i & , \text{if } x_i > \gamma_i \\
\beta_i \times (x_i - \gamma_i) + \zeta_i & , \text{if } x_i \leq \gamma_i
\end{aligned}
\right.
\end{equation}
In RPReLU function $f(\cdot)$, $x_i$ is the input in the $i$th channel, $\gamma_i$ and $\zeta_i$ are learnable shifts, and $\beta_i$ is a learnable coefficient determines the slope of the negative half.

Meanwhile, we also use the default setting that adding parameter-free shortcuts to blocks, similar to~\cite{liu2018bi} and~\cite{liu2020reactnet}. As shown in Figure~\ref{fig:bfblock}, our proposed \textit{Batch Normalization free} (BN-Free) network block maintains the duplication of input activation from~\cite{liu2020reactnet}, replaces by scaling factors (e.g., $\alpha$, $1/\beta_1$, $1/\beta_2$) and adjusts convolutional layers with scaled weight standardization (i.e., WS-Conv).

\paragraph{Distillation loss functions.} To establish the state-of-the-art BNN results, we also introduce the distribution loss function~\cite{liu2020reactnet} to enforce the similarity of distributions between full-precision networks and binary neural networks. It can also be regarded as a knowledge distillation technique. Specifically, the formulation is depicted as follows:

\begin{equation}
\mathcal{L}_{\mathrm{Dis}} = -\frac{1}{n}\sum\limits_c\sum\limits_{i=1}^{n}\rho_c^{\mathcal{R}}(X_i)\times\mathrm{log}(\frac{\rho_c^{\mathcal{B}}(X_i)}{\rho_c^{\mathcal{R}}(X_i)})  
\end{equation}
where $\mathcal{L}_{\mathrm{Dis}}$ is the Kullback–Leibler (KL) divergence, $X_i$ is the input image, $c$ represents classes and $n$ denotes the batch size. $\rho_c^{\mathcal{R}}$ is the softmax output of the full-precision (i.e., real-valued) model and $\rho_c^{\mathcal{B}}$ is the softmax output of the binary neural network. With the assistance of introduced distribution loss, BNN is capable of imitating the prediction distribution from full-precision models, leading to a superior performance. In the implementation, the full-precision NFNet~\cite{brock2021characterizing,brock2021agc} is utilized, which is also a BN-Free network.

\section{Experiments}
\subsection{Setup}\label{sec:setup}
We conduct experiments on three binary models, i.e., XNOR-Net~\cite{rastegari2016xnor}, Bi-RealNet~\cite{liu2018bi}, and ReActNet~\cite{liu2020reactnet} with two widely used backbones, i.e., ResNet-18~\cite{he2016deep} and MobileNetV1~\cite{howard2017mobilenets}. Meanwhile, we evaluate their BN-free counterparts and report the performance on three representative classification datasets, i.e., CIFAR-10~\cite{krizhevsky2009learning}, CIFAR-100~\cite{krizhevsky2009learning}, and ILSVRC12 ImageNet~\cite{russakovsky2015imagenet}. 

\paragraph{Implementation details on ImageNet.} We use the ImageNet dataset with $1000$ classes. There are $1,281,167$ images for training and $50,000$ images for validation. Considering the superior performance of ReActNet~\cite{liu2019circulant} on the ImageNet classification task, we apply our BN-Free network design on ReActNet-18 and ReActNet-A, which are the modifications of ResNet-18 and MobileNetv1 respectively. We also adopt the adaptive gradient clipping (AGC)~\cite{brock2021agc} in the back-propagation when training our BN-Free BNNs with the upper bound value set to $0.02$.

When training the model, We follow the original two-step training strategy~\cite{liu2020reactnet}, where we only binarize the activations and train the network from scratch in the first step, then we fine-tune the network with both binary activations and weights in the second step. In both steps, we train the network for $120$ epochs with the Adam optimizer and an initial learning rate of $5\times10^{-4}$, which follows a linear decreasing scheduler to zero. The weight decay is set to $1\times10^{-5}$ for the first step and $0$ for the second.  Besides, the data augmentation method we used in our experiments follows~\cite{howard2017mobilenets}, which contains random cropping, lighting, and random horizontal flipping. The input resolution is $224$ and the top-1 accuracy on the validation set will be reported in the following section.

\paragraph{Implementation details on CIFAR-10 and CIFAR-100.} Both CIFAR-10 and CIFAR-100 contain $50,000$ training images and $10,000$ testing images from $10$ and $100$ classes respectively. To comprehensively investigate the effectiveness of BN-Free networks, We conduct the classification experiments with four binary networks: XNORNet-18, Bi-RealNet-18, ReActNet-18, and ReActNet-A on CIFAR-10 and CIFAR-100. The first three networks all have a modified ResNet-18 backbone while ReActNet-A is constructed on MobileNetv1. We follow the two-step training strategy consistent with the ImageNet experiments and train the network for $256$ epochs in each step. The upper bound of clipping value in AGC is set to $0.001$ by default, according to the grid searching in  Section~\ref{sec:agc}. Additionally, other training hyperparameters remain the same as those in the ImageNet experiments. Differently, we use only random cropping and horizontal flipping for data augmentation.

\subsection{Comparison to State-of-the-art networks}
We begin by investigating the performance of batch normalization free BNN (BN-Free BNNs). For each network, we apply the Scaled Weight Standardization~\cite{brock2021agc} to all convolution layers and replace the basic blocks with our BN-Free blocks after removing all batch normalization (BN) layers. We report the accuracy between its three variants: the baseline network with BN, the network without BN, and the BN-Free network.

\begin{figure}[htb]
    \centering
    \includegraphics[width=1\linewidth]{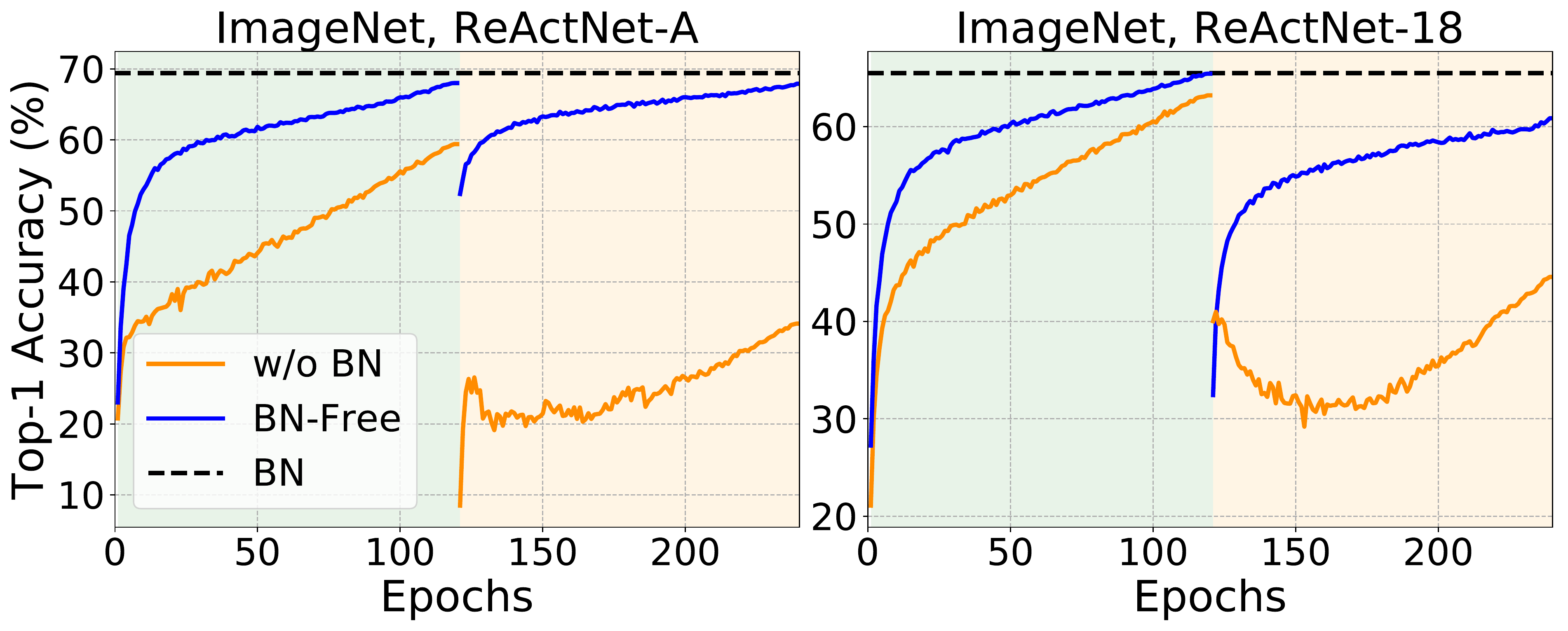}
    \caption{Results of validation accuracy over epochs on ImageNet with ReActNet-18/A. The \textcolor{green}{green} background represents the first training step, in which only activations are binarized. And in the \textcolor{orange}{orange} part, both activations and weights are binary.}
    \label{fig:curve_imagenet}
\end{figure}

\paragraph{Results on ImageNet.} We first evaluate our proposed BN-Free (BF) BNN on ImageNet. Specifically, the BN-Free versions of ReActNet-18 and ReActNet-A are constructed to compare with other existing state-of-the-art BNNs (with BN). Top-1 accuracies are collected in Table~\ref{tab:result_imagenet} and Figure~\ref{fig:curve_imagenet}.

\begin{table}[htb]
\caption{Comparison of the top-1 accuracy with state-of-the-art binary methods on ImageNet. The accuracy of other binary networks are collected from the original papers, which include BNN~\cite{courbariaux2016binarized}, PCNN~\cite{gu2019projection}, XNOR-Net~\cite{rastegari2016xnor}, Bi-RealNet~\cite{liu2018bi}, Real-to-Binary Net~\cite{Martinez2020Training}, ReActNet-18 (BN) and ReActNet-A (BN)~\cite{liu2020reactnet}. ``w/o BN" denotes the version without batch normalization; ``BN-Free" represents our proposed BN-Free BNNs.}
\label{tab:result_imagenet}
\centering
\resizebox{0.40 \textwidth}{!}{
\begin{tabular}{l|c}
\toprule
Binary Network & Top-1 Acc ($\%$) \\\midrule
BNN~\cite{courbariaux2016binarized} & $42.2$ \\
PCNN~\cite{gu2019projection} & $57.3$ \\
XNORNet-18~\cite{rastegari2016xnor} & $51.2$ \\
Bi-RealNet-18~\cite{liu2018bi} & $56.4$ \\ 
Real-to-Binary Net~\cite{Martinez2020Training} & $65.4$ \\ \midrule
ReActNet-18 (BN)~\cite{liu2020reactnet} & $65.5$ \\
ReActNet-18 (w/o BN) & $44.6$ \\
ReActNet-18 (BN-Free) & $61.1$ \\ \midrule
ReActNet-A (BN)~\cite{liu2020reactnet} & $69.4$ \\
ReActNet-A (w/o BN) & $34.1$ \\
ReActNet-A (BN-Free) & $68.0$ \\ 
\bottomrule
\end{tabular}}
\vspace{-4mm}
\end{table}

\begin{figure*}[!htb]
    \centering
    \includegraphics[width=1\linewidth]{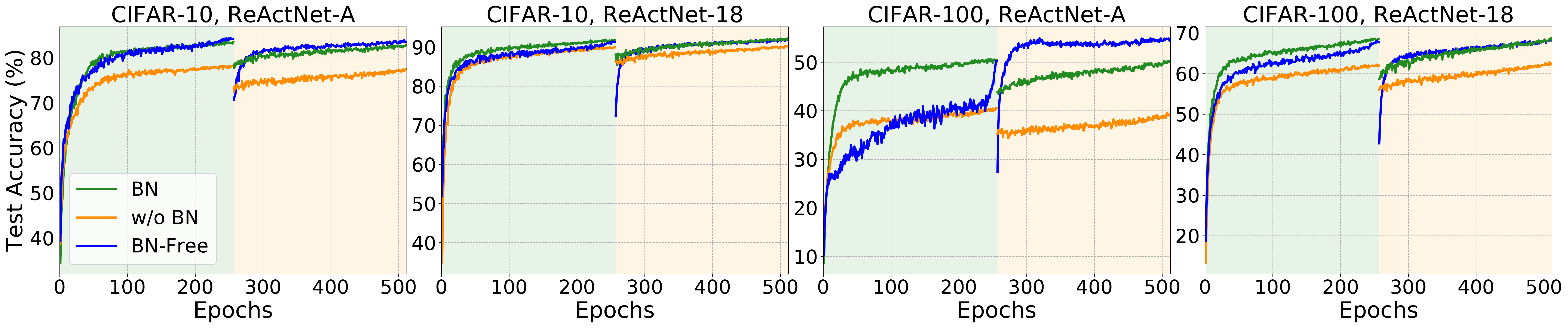}
    \caption{Results of testing accuracy over epochs on CIFAR-10/100 with ReActNet-18/A. The \textcolor{green}{green} background represents the first training step, in which only activations are binarized. And in the \textcolor{orange}{orange} part, both activations and weights are binary.}
    \label{fig:curve_cifar}
\end{figure*}

As shown in Table~\ref{tab:result_imagenet}, compared with the binary neural network without BatchNorm layers, our BN-Free binary neural networks achieve substantial performance improvements. Specifically, we obtain $16.5\%$ and $33.9\%$ accuracy gains for the ReActNet-18 and ReActNet-A on ImageNet, respectively. Note that, our BN-Free ReActNet-A achieves a $68.0\%$ top-1 accuracy, which only has marginal gap (i.e.,$1.4\%$) compared to the state-of-the-art ReActNet~\cite{liu2020reactnet} with batch normalization. Detailed training dynamics are presented in Figure~\ref{fig:curve_imagenet}. We observe that the proposed BN-Free BNN not only reaches a superior performance, but also leads to more stable training. 

\paragraph{Results on CIFAR-10 and CIFAR-100.} To further evaluate the effectiveness of BN-Free modules in BNN, we implement the three binary networks mentioned in section~\ref{sec:setup}, i.e., XNORNet, Bi-RealNet, ReActNet, and compare the performance of their three variants (i.e., BN, w/o BN, BN-Free) on CIFAR-10 and CIFAR-100. With the results in Table~\ref{tab:result_cifar} and Figure~\ref{fig:curve_cifar}, several consistent observations could be drawn as the following:
\begin{itemize}
    \item The proposed BN-Free approach serves as a remedy for the accuracy degradation caused by the absence of BN layers across all datasets and networks. Specifically, when compared with their counterparts without BN, BN-Free BNNs achieve accuracy improvements of $1.75\%\sim8.29\%$, $5.74\%\sim15.63\%$ for different binary networks on CIFAR-10 and CIFAR-100.
    \item Accuracy achieved by BN-Free ReActNet-A surpasses its BN counterpart surprisingly by $0.96\%$ and $4.70\%$ on CIFAR-10 and CIFAR-100, respectively. And BF-ReActNet-18 also achieves comparable performance with its BN version. However, for XNORNet-18 and Bi-RealNet-18, there remains a moderate performance gap between the BN and BN-Free networks. 
    \item Training curves of BN-Free ReActNet on CIFAR-10 in Figure~\ref{fig:curve_cifar}, almost overlaps (BN) ReActNet's curves in both training steps. This indicates our BF networks not only can achieve comparable accuracy but also ensure a stable training process, especially on small datasets.
\end{itemize}

\begin{table}[t]
\caption{\small Comparison of the top-1 accuracy between the three variants (i.e., BN, w/o BN, BN-Free) of binary networks on CIFAR-10 and CIFAR-100. All networks are modified from ResNet-18 except for ReActNet-A, which is constructed from MobileNetv1.}
\vspace{1mm}
\label{tab:result_cifar}
\centering
\resizebox{0.47 \textwidth}{!}{
\begin{tabular}{c|ccc|ccc}
\toprule
\multirow{2}{*}{Binary Network} & \multicolumn{3}{c|}{CIFAR-10 ($\%$)} & \multicolumn{3}{c}{CIFAR-100 ($\%$)} \\ \cmidrule{2-4} \cmidrule{5-7}
& BN & w/o BN & BN-Free & BN & w/o BN & BN-Free \\\midrule
XNORNet-18 & $90.21$ & $71.75$ & $79.67$ & $65.35$ & $45.30$ & $53.76$  \\ \midrule
Bi-RealNet-18 & $89.12$ & $71.30$ & $79.59$ & $63.51$ & $47.72$ & $54.34$ \\ \midrule
ReActNet-18 & $92.31$ & $90.33$ & $92.08$ & $68.78$ & $62.60$ & $68.34$ \\ \midrule
ReActNet-A & $82.95$ & $77.60$ & $\textbf{83.91}$ & $50.30$ & $39.37$ & $\textbf{55.00}$ \\ 
\bottomrule
\end{tabular}}
\end{table}

\begin{table}[t]
\caption{\small Ablation Study of clipping threshold values in AGC on CIFAR-10/100 with ReActNet-18 and ReActNet-A.}
\vspace{1mm}
\label{tab:ablation_agc_cifar100}
\centering
\resizebox{0.47\textwidth}{!}{
\begin{tabular}{c|cc|cc}
\toprule
\multirow{2}{*}{Clipping Value} & \multicolumn{2}{c|}{ReActNet-18} & \multicolumn{2}{c}{ReActNet-A} \\ \cmidrule{2-3} \cmidrule{4-5}
& CIFAR-10 & CIFAR-100 & CIFAR-10 & CIFAR-100\\\midrule
w/o AGC & $91.08$ & $66.15$ & $82.39$ & $48.12$  \\ \midrule
$1\times 10^{-2}$ & $91.23$ & $65.38$ & $82.61$ & $49.46$  \\ \midrule
$5\times 10^{-3}$ & $91.03$ & $66.03$ & $83.10$ & $48.69$  \\ \midrule
$1\times 10^{-3}$ & $\textbf{92.08}$ & $\textbf{68.34}$ & $\textbf{83.91}$ & $51.32$ \\ \midrule
$8\times 10^{-4}$ & $91.54$ & $67.95$ & $83.58$ & $52.27$ \\ \midrule
$5\times 10^{-4}$ & $91.39$ & $67.36$ & $83.31$ & $53.60$ \\ \midrule
$2\times 10^{-4}$ & $90.43$ & $67.07$ & $83.03$ & $\textbf{55.00}$ \\ \midrule
$1\times 10^{-4}$ & $89.62$ & $62.45$ & $80.81$ & $52.28$ \\ 
\bottomrule
\end{tabular}}
\end{table}

\subsection{Ablation Study}
In the previous section, we empirically evaluate the effectiveness of BN-Free modules and verify that our BF-ReActNet-A can reach competitive state-of-the-art performance.
To further investigate the effects of different clipping thresholds in AGC strategy and different components in the proposed BN-Free framework, we provide an ablation study on CIFAR-10 and CIFAR-100 with ReActNet-18 and ReActNet-A as the backbone BNNs.

\begin{figure*}[t]
    \centering
    \includegraphics[width=1\linewidth]{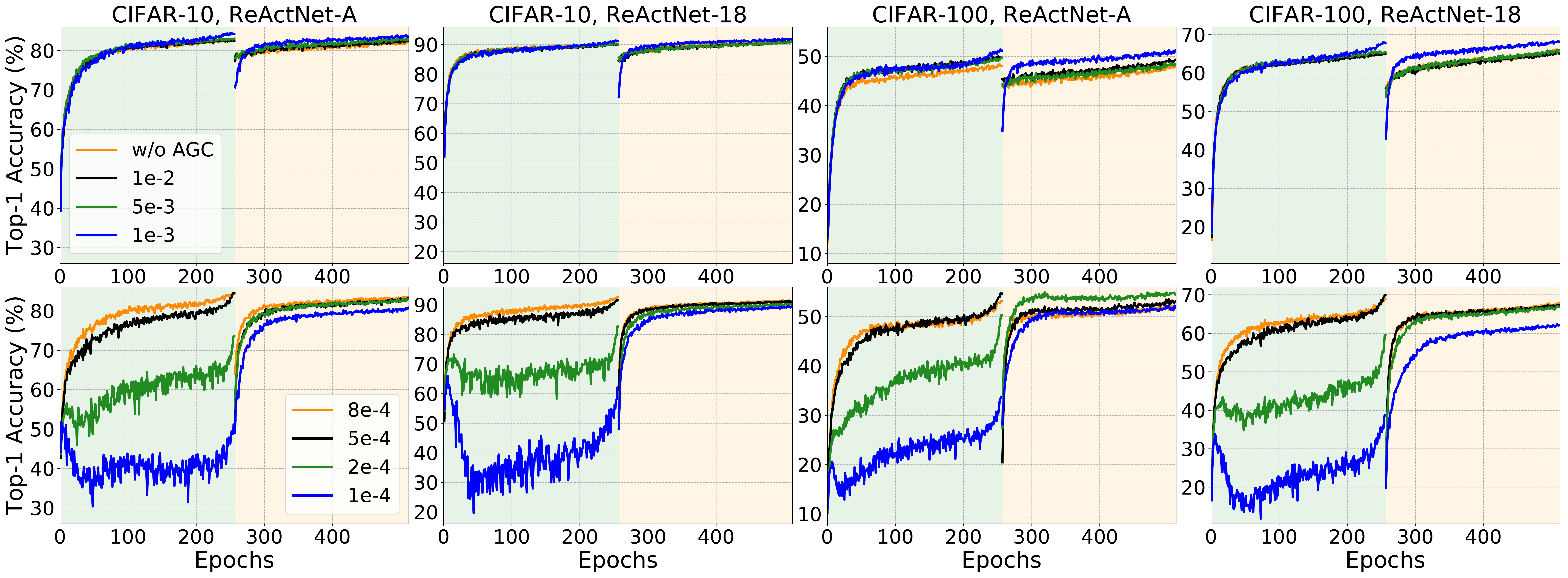}
    \caption{Results of testing accuracy curves of different clipping values in AGC on CIFAR-10/100 with ReActNet-18/A. The \textcolor{green}{green} background represents the first training step, where only activations are binarized. In the \textcolor{orange}{orange} part, both activations and weights are binary. }
    \label{fig:curve_agc}
\end{figure*}

\paragraph{Clipping threshold in AGC.} \label{sec:agc} The clipping threshold $\lambda$ plays an important role in the effectiveness of AGC~\cite{brock2021agc}. In this paragraph, we empirically analyze how does the threshold values affect the training process and final performance. As shown in Table~\ref{tab:ablation_agc_cifar100}, the results indicate that with the growth of the clipping threshold, the final test accuracy first increases to the peak then begins to decline. The results also show that we can get an extra accuracy improvement of $1.00\% \sim 6.88\%$ by using an appropriate threshold. In addition, the performance of BN-Free networks on CIFAR-10 is less affected by the clipping values. A possible explanation is that the performance on the simple CIFAR-10 classification is saturated and less sensitive. Furthermore, Figure~\ref{fig:curve_agc} demonstrates that the training process becomes less stable when the threshold in AGC is extremely small. It comes as no surprise that aggressive gradient clipping introduces undesired noise and causes instability.

\begin{table}[t]
\caption{Ablation study of the separate effect of scaled weight standardization and normalizer-free block on CIFAR-10 and CIFAR-100 with two binary networks based on ReActNet. Test accuracies are reported.}
\label{tab:ablation_nf_cifar}
\centering
\resizebox{0.47 \textwidth}{!}{
\begin{tabular}{c|cc}
\toprule
\multirow{2}{*}{Settings} & \multicolumn{2}{c}{ReActNet-18 ($\%$)} \\ \cmidrule{2-3}
& CIFAR-10 & CIFAR-100 \\ \midrule
BN & $92.31$ & $68.79$ \\
w/o BN & $90.33$ & $62.60$\\ \midrule
WS-Conv & $91.91$ & $68.20$ \\
Specialized Block (i.e., $\alpha,\frac{1}{\beta}$) & $91.44$ & $63.63$ \\ \midrule
BN-Free & $92.08$ & $68.34$ \\ \midrule
\multirow{2}{*}{Settings}  & \multicolumn{2}{c}{ReActNet-A ($\%$)} \\ \cmidrule{2-3}
& CIFAR-10 & CIFAR-100 \\ \midrule
BN & $82.95$ & $50.30$\\
w/o BN & $77.60$ & $39.37$ \\ \midrule
WS-Conv & $82.34$ & $52.37$ \\
Specialized Block (i.e., $\alpha,\frac{1}{\beta}$) & $80.45$ & $54.44$ \\ \midrule
BN-Free & $\textbf{83.91}$ & $\textbf{55.00}$ \\
\bottomrule
\end{tabular}}
\end{table}

\begin{figure*}[t]
    \centering
    \includegraphics[width=1\linewidth]{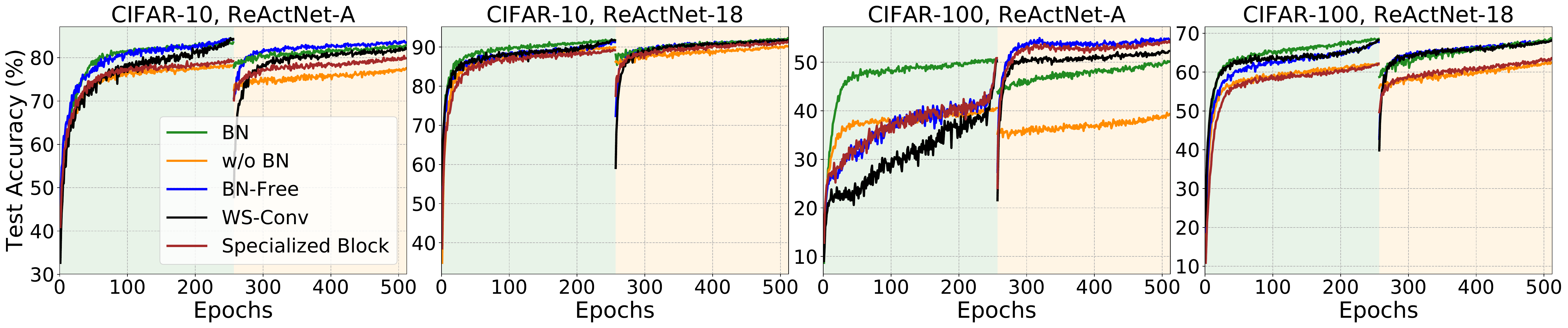}
    \caption{Results of testing accuracy over epochs on CIFAR-10/100 with ReActNet-18/A. The \textcolor{green}{green} background represents the first training step, in which only activations are binarized. And in the \textcolor{orange}{orange} part, both activations and weights are binary. }
    \label{fig:ablation_cifar}
    \vspace{-1mm}
\end{figure*}

\paragraph{Different components in the BN-Free framework.} As described in Section~\ref{sec:method_nf}, the BN-Free module is constructed with a specialized block for the variance normalization, and a scaled weight standardization technique that is applied to all convolution layers (WS-Conv). To study the effects of different components in the BN-Free framework, we construct five variants on top of baseline BNNs (ReActNet-18 and ReActNet-A): \textbf{a)} original baseline (with BN); \textbf{b)} baseline (w/o BN); \textbf{c)} baseline (w/o BN) + WS-Conv; \textbf{d)} baseline (w/o BN) + specialized block; \textbf{e)} baseline (w/o BN) + WS-Conv + specialized block \textbf{which is equivalent to the complete BN-Free setup}. AGC with the best clipping threshold is adopted. The results are collected in Table~\ref{tab:ablation_nf_cifar} and their corresponding training dynamics are presented in Figure~\ref{fig:ablation_cifar}, from which several observations could be drawn:
\begin{itemize}
    \item Either specialized block or WS-Conv can improve the performance of binary networks independently, specifically, the separate improvement achieved by WS-Conv ranges from $1.58\%$ to $13.00\%$ and the separate improvement of the specialized block ranges from $1.03\%$ to $15.07\%$. In addition, the combination of these two approaches can further benefit the BN-Free binary neural networks.
    \item WS-Conv benefits more than the specialized bottleneck by $0.47\% \sim 4.57\%$ performance gains, except the experiment of ReActNet-A on CIFAR-100.
\end{itemize}

\subsection{Visualization}
In this section, we provide the visualization of gradient, latent weight, and activation distributions. Three variants of ReActNet-A (i.e., BN, w/o BN, BN-Free) trained on CIFAR-10 are considered.

\begin{figure}[htb]
    \centering
    \includegraphics[width=1\linewidth]{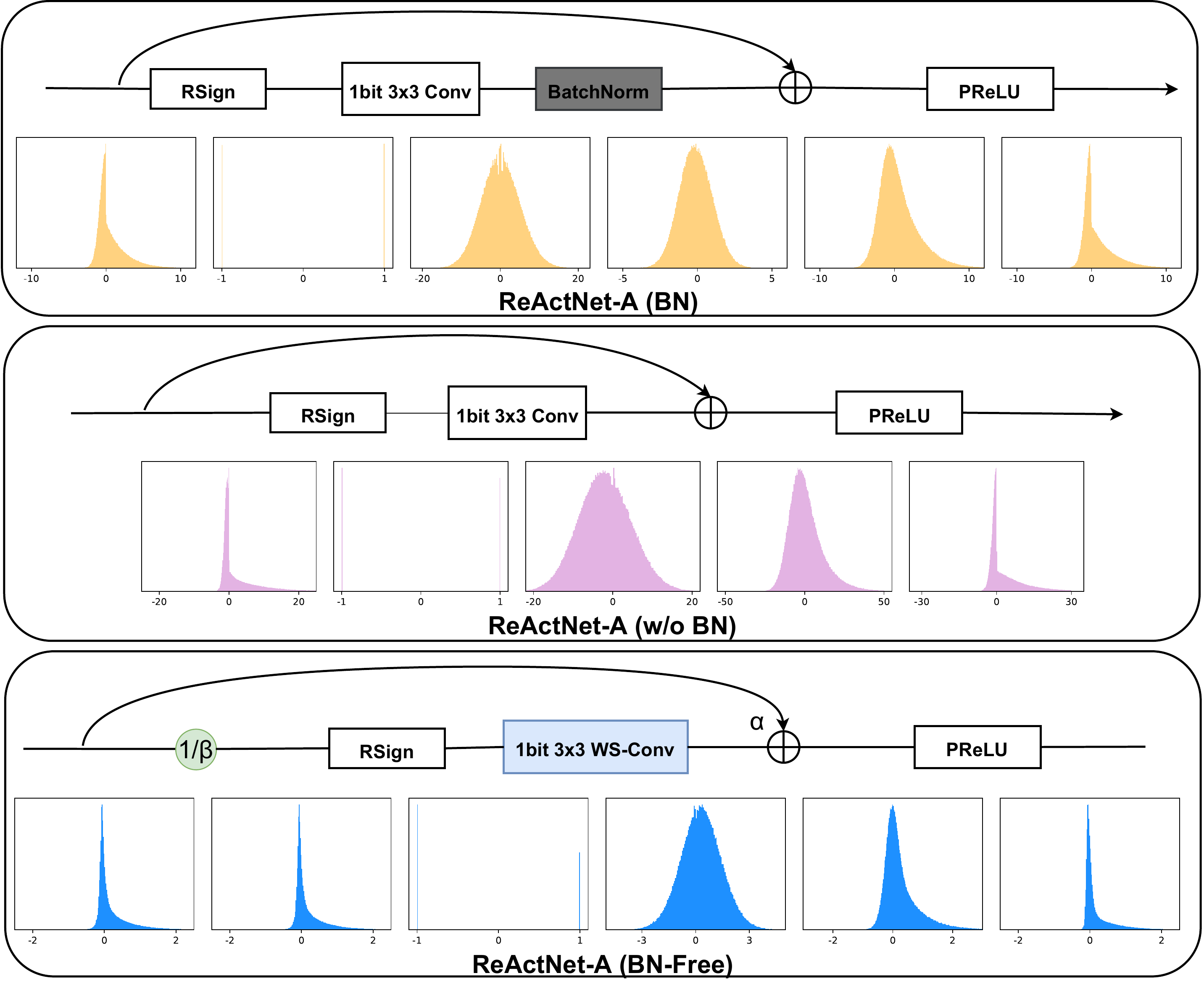}
    \caption{Histogram of the activation distribution inside three variants of ReActNet-A on CIFAR-10: with BN (\textit{top}), without BN (\textit{middle}) and BN-Free (\textit{bottom}).}
    \label{fig:act_dis}
   \vspace{-2mm}
\end{figure}

\paragraph{Activation distribution.} We visualize the activation distribution in Figure~\ref{fig:act_dis}. Compared with the network without BN, the values of the activation inside the BN-Free network are consistently concentrated in a smaller region, which provides some insights into the training stability.

\begin{figure}[htb]
    \centering
    \includegraphics[width=0.8\linewidth]{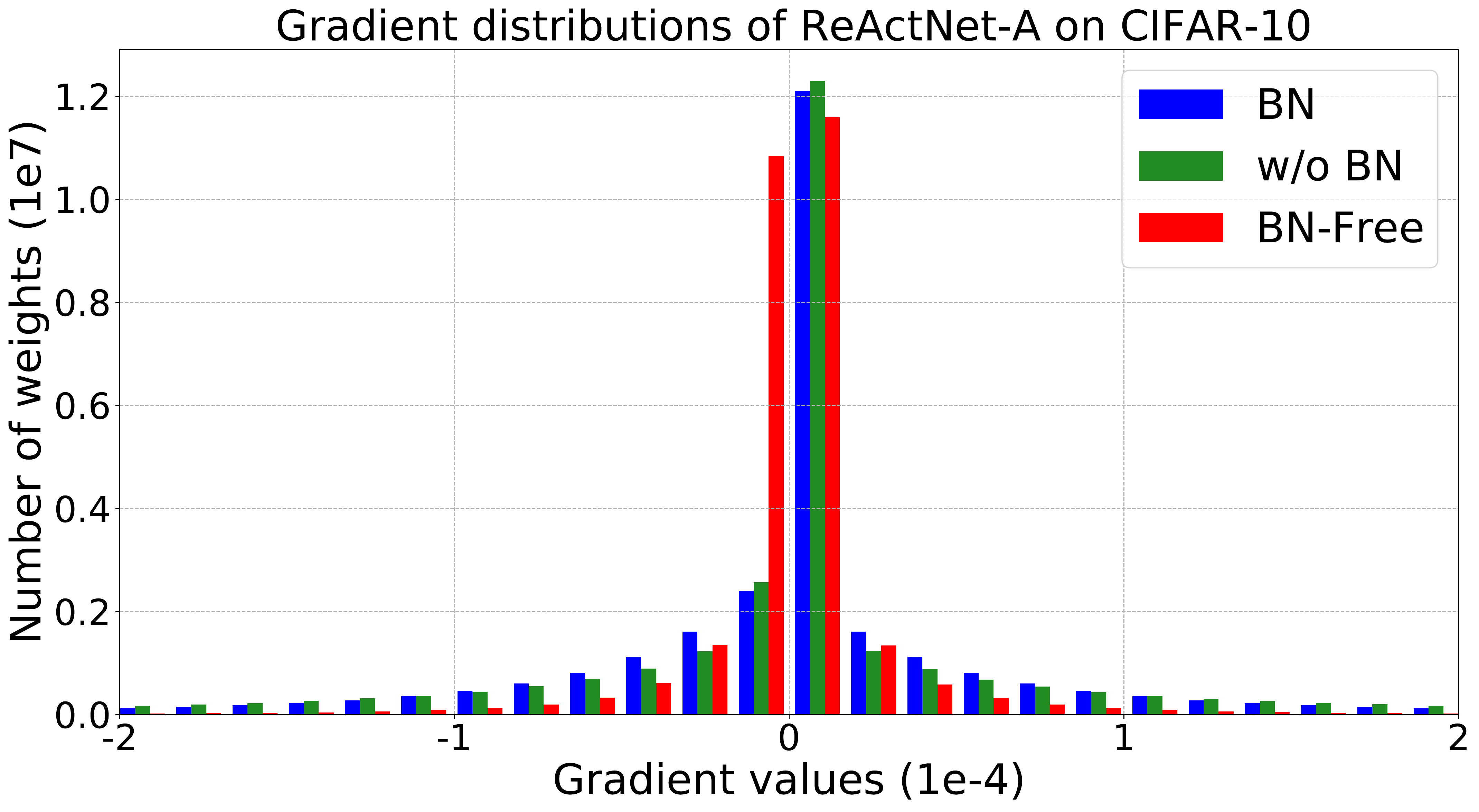}
    \caption{Visualization of gradient distributions of the three variants of ReActNet-A on CIFAR-10: with BN (\textcolor{blue}{blue}), without BN (\textcolor{green}{green}) and BN-Free (\textcolor{red}{red}).}
    \label{fig:grad}
  \vspace{-3mm}
\end{figure}

\paragraph{Gradient distribution.} In figure~\ref{fig:grad}, we show histogram visualizations of the gradient distribution. Our proposed BN-Free BNNs (\textcolor{red}{red} bars) tend to have a smaller range for gradients, which potentially prevents the emergence of gradient exploration caused by training without batch normalization~\cite{santurkar2018does}.

\begin{figure}[t]
    \centering
    \includegraphics[width=0.8\linewidth]{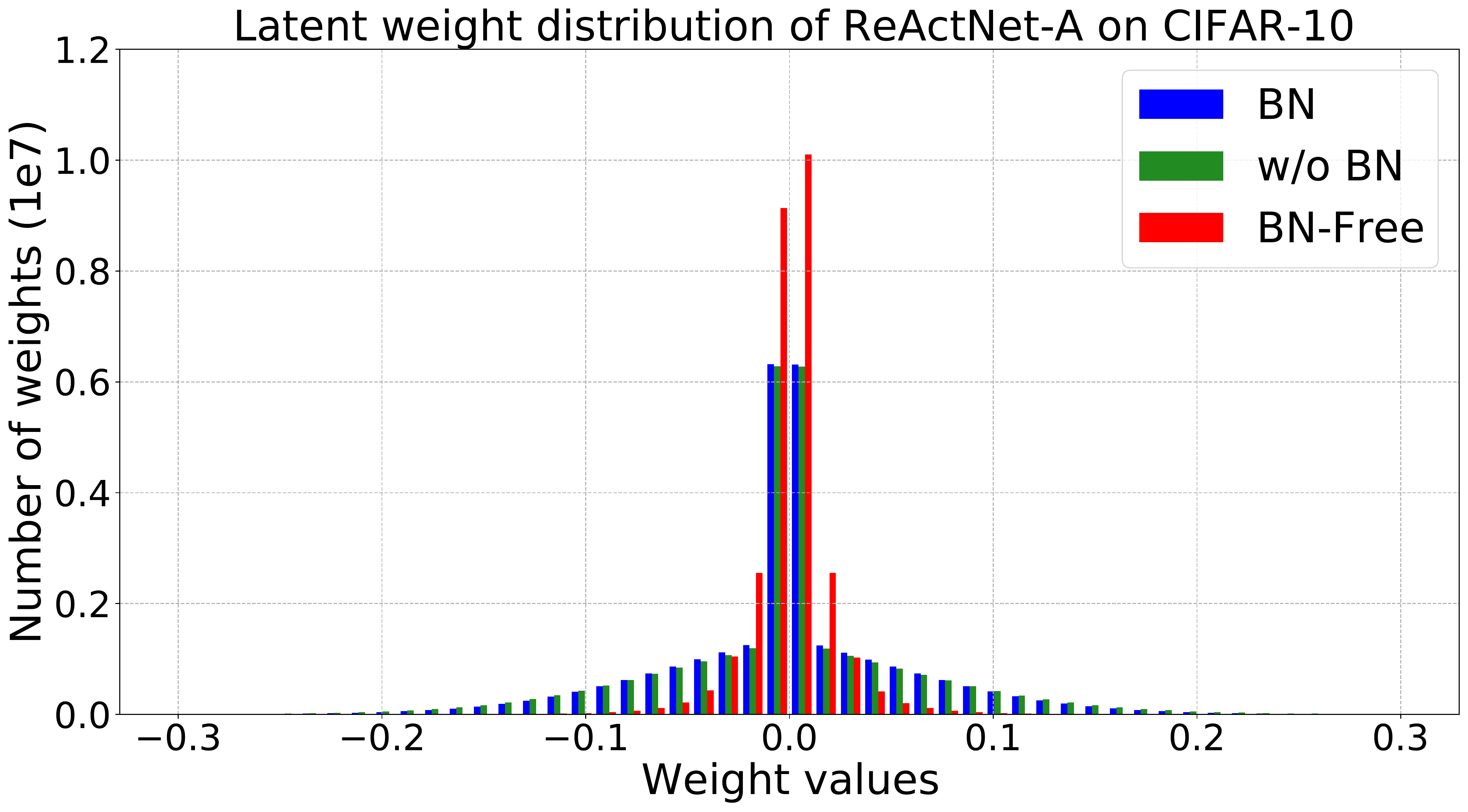}
    \caption{Visualization of latent weight distributions of the three variants of ReActNet-A on CIFAR-10: with BN (\textcolor{blue}{blue}), without BN (\textcolor{green}{green}) and BN-Free (\textcolor{red}{red}).}
    \label{fig:weight}
    \vspace{-1.5mm}
\end{figure}

\paragraph{Latent weight distribution.} Figure~\ref{fig:weight} present the latent weight distribution of three variants of ReActNet-A. We observe that BN-Free BNNs have a more zero-centralized weight distribution, which mainly stems from the weight standardization process.  

\section{Conclusions}
In this paper, we for the first time propose a framework for training binary neural networks without batch normalization, i.e., BN-Free BNN, which achieves competitive state-of-the-art performance compared to its BN-based counterpart. Specifically, We introduce the scaled weight standardization to deal with the mean-shift in the hidden activation distributions caused by removing BN and apply a specialized bottleneck block for the purpose of variance preserving. Moreover, adaptive gradient clipping is adopted to mitigate the gradient exploration issue and stabilize training, for the BN-Free BNN. With the contributions jointly achieved by these techniques, our BN-Free ReActNet achieves $92.08\%$, $68.34\%$, and $68.00\%$ on CIFAR-10, CIFAR-100, and ImageNet, respectively. Note that our BN-Free BNN totally gets rid of batch normalization in both training and inference regimes. In the future, we would be interested to examine the speedup and energy-saving results of the BNN training/inference on a hardware platform.

\clearpage

{\small
\bibliographystyle{ieee_fullname}
\bibliography{bnnobn}
}

\end{document}